# Completing Knowledge by Competing Hierarchies

Kerstin Schill
Institut f. Med. Psychologie
Ludwig-Maximilians-Universität
Goethestr. 31, 8000 München 2
Germany

Ernst Pöppel
Institut f. Med. Psychologie
Ludwig-Maximilians-Universität
Goethestr. 31, 8000 München 2
Germany

Christoph Zetzsche
Lehrstuhl für Nachrichtentechnik
Technische Universität
Arcisstr. 21, 8000 München 2
Germany

## Abstract

A control strategy for expert systems is presented which is based on Shafer's Belief theory and the combination rule of Dempster. In contrast to well known strategies it is not sequentially and hypotheses-driven, but parallel and self-organizing, determined by the concept of information gain. The information gain, calculated as the maximal difference between the actual evidence distribution in the knowledge base and the potential evidence determines each consultation step. Hierarchically structured knowledge is an important representation form and experts even use several hierarchies in parallel for constituting their knowledge. Hence the control strategy is applied to a layered set of distinct hierarchies. Depending on the actual data one of these hierarchies is choosen by the control stratgey for the next step in the reasoning process. Provided the actual data are well matched to the structure of one hierarchy, this hierarchy remains selected for a longer consultation time. If no good match can be achieved, a switch from the actual hierarchy to a competing one will result, very similar to the phenomenon of "restructuring" in problem solving tasks. Up to now the control strategy is restricted to multi-hierarchical knowledge bases with disjunct hierarchies. It is implemented in the expert system IBIG (inference by information gain), being presently applied to acquired speech disorders (aphasia).

## 1 Introduction

The performance of an expert system is essentially determined by two important factors: the completeness of the knowledge base with respect to the expert knowledge and the ability of the control strategy to make adaquate decisions even in those cases where the actual data are irregularly distributed across the knowledge base. To overcome these problems we shall introduce a knowledge base enabling the incorporation of diverse knowledge structures and a formally derived control strategy. The control strategy is based on the concept of information. Information gain is calculated as the difference between the actual evidence distribution and the potential one, where evidence is handled by Shafer's belief theory and Dempster's rule of combination. In the first section of this paper, we shall introduce the multi-level knowledge base. The control strategy will be described in the second section. The strategy is first derived for a single hierarchy and is then extended to a restricted version of the multi-level model, a model with an arbitrary number of disjunctive hierarchies.

## 2 The knowledge base

### 2.1 Common knowledge bases

Common knowledge bases in expert systems with a symbolic knowledge representation can be classified into two categories: single-structure knowledge bases, like the one representing knowledge within one strict hierarchy (e.g., the medical expert system MEDIKS (Chang 1984)), and integrated structures where two or more relationships are linked in one net structure (e.g., the systems EXPERT (Weiss 1979) or CADUCEUS (Miller 1984)). The first category has to deal with the problem that, in most cases, the knowledge of a certain domain is not completely covered by one structure, e.g. one single hierarchy. This implies that necessary inferences cannot be drawn in the reasoning process. Thus, the idea of one best structuring principle for all problems is not suitable (Schill 1986, 1990). The second category has to deal with the problem that no formally derived and transparent method exists to calculate evidence. Since this calculation is not theoretically justified, the same is true for the reasoning process based on it.

Common control strategies for such knowledge bases can be characterized by the principle of partitioning. In rule-based systems a single hypothesis or an agenda with a small number of probable hypotheses are partitioned from all others and control the data collection process (see e.g. MYCIN (Shortliffe 1976)). In hierachically structured knowledge bases a continuous partitioning process starting at the root guides the selection process of the data. The main problem of these partitioning principles arise in unclear consultation cases where the decision has to be taken which of the hypotheses has to be selected from a large list of weak hypotheses. The wrong selection may lead to a "dead end", with no solution or an expensive backtracking.

## 2.2. A multi-hierarchical knowledge base

Expert knowledge can be characterized by two components, i.e. knowledge entities and the relationships between them. Knowledge entities can be hypotheses or sets of hypotheses, for example diseasecategories or diseases themselves. Typical examples for relationships are hierarchical, causal, or temporal ones. Only both components, entities and relationships together, allow the constitution of expert knowledge. Thus, a knowledge-base structure is required which allows the representation of many diverse relationships for linking the knowledge entities to be made.

In the expert system IBIG (inference by information gain) (Schill 1986), a knowledge-base structure is developed that meets the above-mentioned requirements or supports their further development. The knowledge base is modeled by the parallel representation of an arbitrary number of separate layers. The idea is to represent on each level or layer one specific relationship linking the appropriate knowledge entities. Thus, such a base may have one causal layer, perhaps two different hierarchical ones (knowledge of a certain domain can often be represented by two or more different hierarchies), and a layer characterized by temporally linked knowledge, etc..

A formally derived control strategy applicable to a restricted form, namely a layered set of disjunctive hierarchies, has been developed. Modelling expert knowledge, we have observed that such hierachical structuring is characteristic of many problem-solving situations. As a step towards a complete knowledge representation, not one hierarchy but many distinct ones have to be taken into account for the reasoning process. Thus our actual knowledge base consists of an arbitrary number of strict hierarchies $T_i$,
where for all hierarchies $T_i$ and $T_j$, $i \neq j$, $i, j = 1....n$,

if $X_i \subset T_i$ and $X_j \subset T_j$ then $X_i \cap X_j = \emptyset$, for all subsets $X_i$ and $X_j$. The actual restriction to disjunctive hierarchies is necessary in order to guarantee the consistent distribution of belief induced by pieces of evidence.

To derive our control strategy, first we briefly show the basic formulars of Shafer's belief theory and Dempster's combination rule. An approximation of this theory for combining evidence in hierarchies is introduced next. This approach is used to calculate the evidence in IBIG and is a base for the information increment strategy.

## 3. The control strategy

### 3.1 Shafer's Belief theory

Shafer's Belief Theory (Shafer 1976) is based on subjective belief measures induced in experts given some pieces of evidence. The axiom $Bel(A) + Bel(A^c) \leq 1$ enables one to distinguish betweeen lack of knowledge and nonsupporting knowledge. A set of propositions about the mutually exclusive and exhaustive possibilities in a domain is called the frame of discernment and is denoted by $\theta$. It's set of subsets is denoted by $2^\theta$ where elements of $2^\theta$ are the general propositions in the domain. It is postulated that some finite amount of belief can be spread among various propositions A of $\theta$ according to the available evidence, with one and only one true proposition. This quantity of belief m(A) is allocated to the proposition $A \subset \theta$ and is called basic probability number or basic belief mass (Smets 1988) and represents our exact belief in the proposition represented by A.

A function called basic probability assignment or basic belief assignment assigns to each subset of $\theta$ a measure of our belief in the proposition represented by the subset. It is defined whenever

$$m: 2^\theta \rightarrow [0,1], \quad m(\emptyset) = 0 \text{ and } \sum_{A \subset \theta} m(A) = 1 \text{ is satisfied.}$$

In terms of this basic belief assignment, the belief in a proposition $A \subset \theta$ can be expressed as: $Bel(A) = \sum_{B \subset A} m(B)$

Pieces of evidence are combined by the application of Dempster's rule of combination on the basic belief assignments. If two distinct pieces of evidence induce two basic belief masses $m_1$ and $m_2$, the product of these masses is allocated to the conjunction of the two focal propositions $A_i$ and $B_j$. Thus, the combination of two basic belief masses is defined by:

$$m_{12}(A) = K * \sum_{\substack{i,j \\ A_i \cap B_j = A}} m_1(A_i) * m_2(B_j)$$

with the normalization constant K

$$K = (1 - \sum_{\substack{i,j \\ A_i \cap B_j = \emptyset}} m_1(A_i) * m_2(B_j))^{-1}.$$

The effects of this normalization with respect to an open or closed world assumption is critically discussed in (Smets 1989).
Since the above shown formulas are the basic one's for the following derivations for further details see (Shafer 1976).

### 3.2 The Belief theory applied to hierarchies

Based on Barnett's approach (Barnett 1981), Gordon and Shortliffe have developed an approximation of the belief theory applicable to strict hierarchical knowledge structures. Later, a correct approach requiring no essential change of the proposed strategy was derived by Shafer and Logan (Shafer 1987). Since the belief approach is applied to $2^\theta$, the important step of Gordon and Shortliffe for the application to hierarchies was the approximation that pieces of evidence for sets not represented in the tree must be associated with the smallest supersets represented in the tree. The approach is based on simple support functions, a subclass of belief functions for which the concept of confirming and non-confirming belief has to be introduced. In the approach of Gordon and Shortliffe non-confirming belief for a set in the tree is confirmingly allocated to the set-theoretical complement.

In the following, we briefly show the resulting combination rules with which the calculation of the basic belief mass is provided in a three-step technique. For further details see (Gordon 1985).

Given some pieces of evidence which induce belief masses $m_{ij}$ for a set $X_i$ in our hierarchy, we start by combining these masses to reach a combined belief mass pointed exactly at $X_i$ and no other set in the tree. The same has to be provided for non-confirming belief $m_{ij}^c$ and the concerned set $X_i^c$ in T' ( T' is the set of all set-theoretical complements of T). This first step must be applied to all sets in the tree.

Thus, calculate $m_{X_i}$ for all $X_i$ in T and $m_{X_i}^c$ for all $X_i^c$ in T' with:

$$m_{X_i} = 1 - \prod_{j=1}^{n}(1-m_{ij}) \text{ and } m_{X_i}^c = 1 - \prod_{j=1}^{n}(1-m_{ij}^c)$$

In the second step, the combination of confirming belief for each set in the hierarchy with respect to all other sets is provided by calculating the aggregated basic belief mass defined by:

$$m_T(X_i) = K * m_{X_i}(X_i) * \prod_{\substack{X_j \not\supset X_i \\ X_j \in T}} m_{X_j}(\theta) \text{ if } X_i \in T \text{ or}$$

$$m_T(X_i) = K * \prod_{X_j \in T} m_{X_j}(\theta) \text{ if } X_i = \theta$$

In the last step, the combination of nonconfirming belief for all sets in the tree which has attained evidence in any earlier step is provided with the calculation:

$m_T \ominus m_{X_1}^c$, then ($m_T \ominus m_{X_1}^c) \ominus m_{X_2}^c$, etc. for all $X_i^c$ in T'.

Case 1: $X_j \subseteq X_i$ :
  $m_T \ominus m_{X_i}c(X_j) = K * m_T(X_j) * m_{X_i}c(\theta)$.

Case 2: $X_j \cap X_i = \emptyset$
  if $X_j \cup X_i$ is a set in $T \cup \theta$ :
  $m_T \ominus m_{X_i}c(X_j) =$
    $K * [m_T(X_j) + m_T(X_j \cup X_i) * m_{X_i}c(X_i^c)]$
  if $X_j \cup X_i$ is not a set in $T \cup \theta$ :
  $m_T \ominus m_{X_i}c(X_j) = K * m_T(X_j)$.

Case 3: $X_j \supset X_i$
  if $X_j \cap X_i^c$ is not a set in T :
  $m_T \ominus m_{X_i}c(X) = K * m_T(X_j)$
  if $X_j \cap X_i^c$ is a set in T :
  $m_T \ominus m_{X_i}c(X_j) = K * m_T(X_j) * m_{X_i}c(\theta)$.

With these three cases, the calculation of belief masses induced by pieces of evidence is provided in IBIG.

### 3.3 The information increment strategy

The development of our control strategy was guided by two main ideas. First, to make a strategy available which adapts itself to the actually existing data situation and which is not predetermined like most of the known strategies. As mentioned above, these predetermined strategies, characterized by hypotheses-driven behaviour, may culminate in an unsuccessful search whenever data are ambiguous or irregularily distributed over the hierarchy and therefore not represented in one single path. In all these cases, the rigid, top-down partitioning processes are not optimal. The second requirement for our strategy was to make a theory-based strategy available which provides an axiomatic base for further development.

The principles of our strategy first described in (Schill 1986) is to calculate the data one has to collect next to attain the largest information gain with respect to the actual data situation. The information gain is calculated by the difference between the belief distribution in the hierarchy, induced by the available pieces of evidence, and the "potential belief distribution".

Each data item in knowledge base indicating evidence for a set of hypotheses is associated with an a priori belief mass. The potential belief mass is identical with this a priori measure as long as nothing is known about this data item. The following derivations show the influence of all these potential belief masses in the hierarchy on the actual given belief situation. This has to be derived for two extreme cases, namely, the potential confirming belief and the non-confirming belief for these data. In the following, the potential belief of a node or hypotheses set $X_i$ in the tree is designated by $\hat{m}_i$.

In the first case, we assume that all presently unknown or uncollected data, representing evidence which induce confirming belief for a set of hypotheses $X_i$ will be collected and confirmed. Thus, corresponding to Gordon and Shortliffe's first step, the potential belief is calculated as follows:

$$\hat{m}_{X_i}(X_i) = 1 - \prod_{j=1}^{n}(1-\hat{m}_{ij}), \text{ where } \hat{m}_{ij} \text{ are a priori given}$$

maximal potential belief masses.

In the second case, we analogously combine the belief for the potential non-confirming case:

$\hat{m}_{x_i}^c(X_i) = 1 - \prod_{j=1}^{n}(1-\hat{m}_{ij}^c)$, where $\hat{m}_{ij}^c$ are a priori given maximal non-confirming belief masses

Up to now, we have calculated the potential belief masses $(\hat{m}_{x_j}(X_j), \hat{m}_{x_j}^c(X_j))$ a node or hypothesis set $X_j$ may reach with respect to all presently unknown data.

The next step is to analyze the influences of this potential belief on the actual belief distribution. For this analysis, we examine elementary evidence situations in the hierarchy consisting of only one node or hypothesis set $X_i$ with actual belief. It is assumed that each more complex belief situation can be approximated by the superposition of elementary evidence situations. Using the approximation of Gordon and Shortliffe, we now calculate the influence of nodes $X_j$ in the tree on the hypothesis set $X_i$. This is expressed by the potential belief $\hat{m}_T(X_i)$, which $X_i$ attains from $X_j$ proportional to its potential belief mass $\hat{m}_{x_j}$. The information increment for $X_i$ arising from $X_j$ is calculated next by the difference between the belief mass $m_T(X_i)$ and the potential value $\hat{m}_T(X_i)$.

Depending on the intersection possibilities in the hierarchy, the calculation of the information increment is derived in five equations. These five equations include all possible intersections and influences between any node $X_j$ and $X_i$ in the tree. Thus, the information-increment calculation is computed in the following way: On every node or hypothesis set $X_i$ in the hierarchy which has attained evidence, the equations are applied. The information-increment arising from a node $X_j$ is allocated to this node. If more than one equation is applied to $X_j$, the information increment is added. This superposition has been used to obtain an information increment measure related to the complete evidence distribution. The node or hypothesis set with the largest information gain is the one which determines the data-collection process in the next step. For an extensive derivation of the following five equations see (Schill 1986).

The first equation describes the information increment for the node $X_i$ from $X_j$ with $X_j \not\supset X_i$. It is the difference between the belief mass $m_T(X_i)$ and the potential belief mass $\hat{m}_T(X_i) = (K * m_{x_i}(X_i) * \prod_{\substack{X_j \not\supset X_i \\ X_j \in T}} \hat{m}_{x_j}(\theta))$ induced by

he potential confirming belief of $X_j$. Thus it is calculated by:

Equation 1: $X_j \not\supset X_i$
(all nodes without father nodes and the node itself have potential confirming belief)

$I_{wFC} = | m_T(X_i) - (K * m_{x_i}(X_i) * \prod_{\substack{X_j \not\supset X_i \\ X_j \in T}} \hat{m}_{x_j}(\theta)) |$

The four other equations are analogously derived.

Equation 2: $X_j \cap X_i = \emptyset$, $X_j \cup X_i$ is in T ( binary siblings which have potential nonconfirming belief)

$I_{SIN} = $
$| m_T(X_i) - (K * [m_T(X_i) + m_T(X_i \cup X_j) * \hat{m}_{x_j}^c(X_j^c)]) |$

Equation 3: $X_j \supset X_i$ (the node itself or the father nodes have potential nonconfirming belief)

$I_{IFN} = | m_T(X_i) - (K * m_T(X_i) * \hat{m}_{x_j}^c(\theta)) |$

Equation 4: $X_j \subset X_i$, $X_j \cap X_i^c \in T$ ( $X_i$ is in a binary sibling situation and also the son nodes $X_j$, which have potential non-confirming belief)

$I_{SN} = | m_T(X_i) - (K * m_T(X_i) * \hat{m}_{x_j}^c(\theta)) |$

Equation 5: $X_j = X_i$ ( the node itself has still potential confirming belief)

$I_{NC} = | m_T(X_i) - (K * \hat{m}_{x_i}(X_i) * \prod_{\substack{X_k \not\supset X_i \\ X_k \in T}} m_{x_k}(\theta)) |$

The suggested information increment, measures the potential change of the knowledge state which can be caused by new, incoming data.
This use of the concept of information is basically comparable to similar recent approaches (Klir 1987, Smets 1983).

## 4. The application to a multi-hierarchical knowledge base

For a single hierarchy the information increment strategy described above, results in many different adaptive behaviours. Thus e.g. depending on the actual data situation, the control strategy is organizing itself as a parallel search where around the centers of evidence the node that promises the largest information gain is choosen. Another extreme behaviour arises if all data are strictly lying on one single path, in which case the well known top-down partitioning behaviour can be observed.

Given a knowledge base including a number of distinct hierarchies and some preliminary evidence, the information increment strategy calculates in parallel the belief distribution for every hierarchy separately. In the next step, the strategy computes the information gains for each level separately. The hypothesis set with the maximal information gain with respect to all levels determines the next consultation step. The unknown or uncollected data of this node will be collected and, depending on their coincidence, the belief situation is recalculated at all levels in the knowledge base. The control strategy thus applied to the multi-hierarchy structure is an extension of the previous one with one hierarchy. Depending on the actual consultation, the control strategy might operate at one level for the whole reasoning process. This appears whenever the structure of the problem-solving task, e.g. the structure of the disease and the available symptoms, fits optimally into the format of this hierarchical structure. In the extreme opposite case, the information-increment strategy shifts from one hierarchy in the knowledge base to another. New data collected by the largest information increment at one hierarchy allow the strategy to switch to another hierarchy. Cognitively speaking, these switches can be considered as the restructuring of the problem-solving task.

## 5. Discussion

A knowledge representation scheme enabling the parallel representation of various knowledge relationships has been developed. This type of representation can be regarded as a first step towards the development of knowledge base structures which can exhaustively cover the variety of structural relationships found in knowledge of human experts. In contrast to other approaches the clear separation of the knowledge structures ensures easy consistency checks and knowledge acquisition.

For the restriction to distinct hierarchies, a formally based control strategy was developed which is adaptive with respect to the actual data available. For this control strategy, based on the concept of information, no predetermined direction in the form of a "leading hypothesis" is given. Thus dead-end behaviour arising from the successive hypotheses-driven strategies can be avoided. Depending on the structure of the problem-solving task, the strategy operates by choosing one hierarchy for a longer consultation time or by switching between competing hierarchies, thereby implicitly restructuring the problem-solving task. The calculation of evidence, which is an important basis for the strategy, is provided by an approximation of Shafer's belief theory and Dempster's combination rule. This theory allows the use of subjective belief measures of experts and enables one to operate with uncertain knowledge in an adaquate way.

The application has thus far been restricted to knowledge structures in the form of distinct hierarchies in order to guarantee the consistent distribution of belief in the knowledge base. We have implemented this multi-level representation scheme, applying the information-increment strategy in the expert system IBIG which is running on a SUN-3 workstation. With a hypothetical knowledge base consisting of 8 different hierarchies, where each hierarchy represents ca. 300 nodes, the running time for evaluating the largest information gain is 29 seconds. For a "real" medical application the system has already been successfully tested in a small number of cases. Application to a medical area with a larger knowledge base then the one of acquired speech disorders is currently in progress.

### Acknowledgements

I thank Christian Freksa and Till Roenneberg for theirs support.